\documentclass[letterpaper,10pt,conference]{ieeeconf}
\IEEEoverridecommandlockouts

\makeatletter

\usepackage{amsmath,amsfonts,amssymb}

\let\labelindent\relax
\usepackage{prettyref}
\usepackage{mathrsfs}
\usepackage{graphicx}
\usepackage{wrapfig}
\usepackage{subfig}
\usepackage{bbold}
\usepackage{tabu}
\usepackage{MnSymbol}
\usepackage{multirow}
\usepackage{booktabs}
\usepackage{enumitem}
\usepackage{algorithm}
\usepackage[table]{xcolor}
\usepackage[noend]{algpseudocode}
\usepackage
[backend=bibtex,
bibstyle=ieee,
citestyle=numeric,sortcites,
mincitenames=1,
maxcitenames=2,
natbib=true,
doi=false,
isbn=false,
url=false,
eprint=false]{biblatex}
\usepackage[colorlinks,allcolors=gray,hypertexnames=true]{hyperref} 
\newcommand{\citewithauthor}[1]{\citeauthor{#1} \cite{#1}}

\algnewcommand{\IfThenElse}[3]{
  \State \algorithmicif\ #1\ \algorithmicthen\ #2\ \algorithmicelse\ #3}
  
\algnewcommand{\LineComment}[1]{\State \(\triangleright\) #1}
\algdef{SE}[DOWHILE]{Do}{doWhile}{\algorithmicdo}[1]{\algorithmicwhile\ #1}
\newcommand*{\colorboxed}{}
\def\colorboxed#1#{%
  \colorboxedAux{#1}%
}
\newcommand*{\colorboxedAux}[3]{%
  \begingroup
    \colorlet{cb@saved}{.}%
    \color#1{#2}%
    \boxed{%
      \color{cb@saved}%
      #3%
    }%
  \endgroup
}

\newrefformat{fig}{Figure~\ref{#1}}
\newrefformat{par}{Section~\ref{#1}}
\newrefformat{appen}{Appendix~\ref{#1}}
\newrefformat{sec}{Section~\ref{#1}}
\newrefformat{sub}{Section~\ref{#1}}
\newrefformat{table}{Table~\ref{#1}}
\newrefformat{ass}{Assumption~\ref{#1}}
\newrefformat{alg}{Algorithm~\ref{#1}}
\newrefformat{def}{Definition~\ref{#1}}
\newrefformat{thm}{Theorem~\ref{#1}}
\newrefformat{cor}{Corollary~\ref{#1}}
\newrefformat{lem}{Lemma~\ref{#1}}
\newrefformat{step}{Step~\ref{#1}}
\newrefformat{ln}{Line~\ref{#1}}
\newrefformat{rem}{Remark~\ref{#1}}
\newrefformat{eq}{Equation~\eqref{#1}}
\newrefformat{pb}{Problem~\ref{#1}}
\newrefformat{it}{Item~\ref{#1}}
\newrefformat{te}{Term~\ref{#1}}
\def\Eqref Eq:#1:{\eqref{eq:#1}}
\newrefformat{Eq}{Equation~\Eqref#1:}

\newcommand{\TE}[1]{\textbf{#1}}

\newcommand{\argmin}[1]{\underset{#1}{\text{argmin}}\;}

\newcommand{\ST}{\text{s.t.}\quad}

\usepackage{xcolor}
\definecolor{Blue} {rgb}{0.4, 0.4, 0.8}
\definecolor{Red}  {rgb}{0.8, 0.2, 0.2}
\definecolor{Green}{rgb}{0.2, 0.8, 0.2}

\usepackage{xcolor}

\newif\ifarxiv
\arxivtrue

\usepackage[margin=0.75in]{geometry}

\def\BibTeX{{\rm B\kern-.05em{\sc i\kern-.025em b}\kern-.08em
    T\kern-.1667em\lower.7ex\hbox{E}\kern-.125emX}}

\begin{document}
\allowdisplaybreaks

\title{\LARGE \bf Language-Guided Generation for Personalized Inspection Planning}
\author{Xingpeng Sun$^1$, Zherong Pan$^2$, Xifeng Gao$^2$, Kui Wu$^2$, Aniket Bera$^1$ \thanks{$^1$ Purdue University, USA. $^2$ LightSpeed Studios, USA.}}

\maketitle

\begin{abstract}
We propose a training-free, Vision-Language Model (VLM)-guided approach for efficiently generating trajectories to facilitate target inspection planning based on text descriptions. Unlike existing Vision-and-Language Navigation (VLN) methods designed for general agents in unknown environments, our approach specifically targets the efficient inspection of known scenes, with widespread applications in fields such as medical, marine, and civil engineering. Leveraging VLMs, our method first extracts points of interest (POIs) from the text description, then identifies a set of waypoints from which POIs are both salient and align with the spatial constraints defined in the prompt. Next, we interact with the VLM to iteratively refine the trajectory, preserving the visibility and prominence of the POIs. Further, we solve a Traveling Salesman Problem (TSP) to find the most efficient visitation order that satisfies the order constraint implied in the text description. Finally, we apply trajectory optimization to generate smooth, executable inspection paths for aerial and underwater vehicles. We have evaluated our method across a series of both handcrafted and real-world scanned environments. The results demonstrate that our approach effectively generates inspection planning trajectories that adhere to user instructions.
\end{abstract}

\section{Introduction}
Inspection planning is a crucial component of autonomous robotic systems tasked with monitoring and assessing known environments. This problem has a wide range of applications, including fault detection in mechanical parts~\cite{raffaeli2013off}, underwater surveying of shipwrecks~\cite{bingham2010robotic}, and structural inspections in civil engineering~\cite{bircher2015structural}. In these scenarios, the objective is for an aerial or ground vehicle to efficiently inspect a set of points of interest (POIs). Over the past decade, various methods for coverage and inspection planning have been proposed~\cite{ayvali2017ergodic,fu2021computationally,8338068}. However, POIs in inspection planning typically need to be manually defined, either as waypoints or topological constraints, requiring domain-specific knowledge. This dependence can limit the accessibility and adaptability of these methods for non-expert users or in complex environments.
Recently, learning-assisted trajectory generation has emerged as a promising solution. These methods automatically convert high-level instructions into low-level inputs, with notable approaches including vision-based target recognition and tracking algorithms~\cite{wang2020smart,ryll2020semantic}. While these algorithms can autonomously generate tracking or goal-reaching trajectories, they often rely on domain-specific vision models. Adapting these models to new domains typically requires data-intensive retraining or a complete algorithm redesign. Some vision models, such as YOLO~\cite{redmon2016you}, can perform open-domain object detection from a single image, but they lack the ability to reason across multiple views or infer spatial relationships between camera poses and the observed content. This limitation highlights the need for intuitive, high-level interfaces that bridge human intent with low-level trajectory generation. Recent advancements in open-domain vision-language models (VLMs) offer a promising solution, as they can ground natural language descriptions in visual context, enabling robots to generate inspection trajectories directly from user-provided text prompts.
\begin{figure}[t]
\centering
\includegraphics[width=\columnwidth]{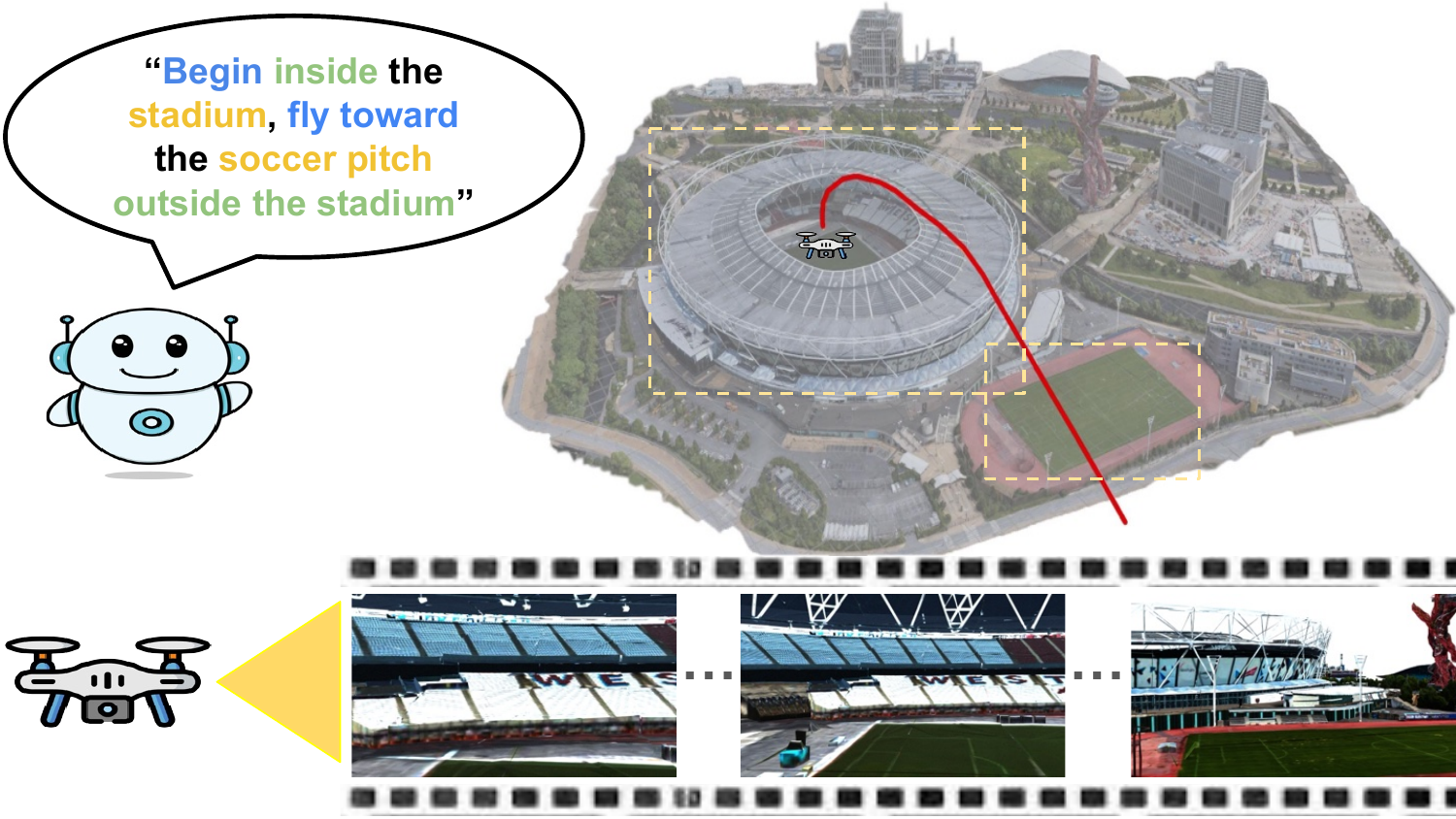}
\caption{\label{fig:teaser}\footnotesize{Our method generates target inspection planning trajectory (red) that conforms to the text description (top left). The text description contains a set of POIs (dashed yellow boxes), the relative robot poses required to observe these POIs (green), the desired visitation order of POIs (blue), and the bottom images show these locations in the first person.}}
\vskip -0.3in
\end{figure}

The latest success of Vision-Language Models (VLMs) has sparked significant interest in using text and visual inputs to guide robotic trajectory generation. Specifically, the Vision-and-Language Navigation (VLN) paradigm enables agents to explore unknown environments by following intuitive, high-level instructions~\cite{gu2022vision}. A key strength of pre-trained VLMs is their strong reasoning capabilities and ability to interpret navigation commands without requiring task-specific fine-tuning~\cite{zhou2024navgpt, sun2024trustnavgpt}, thus significantly reducing the need for data-intensive retraining or complete algorithm redesigns.
However, while VLN holds promise, these methods typically assume that the agent is environment-agnostic and needs to identify POIs through exploration. In contrast, inspection planning involves environments that are already known, with the primary goal being for the agent to reach a set of predefined POIs as efficiently as possible. Inspection is often performed by aerial or underwater vehicles with limited battery life, further emphasizing the need for efficient trajectory planning.
A recent attempt to adapt VLN for UAV trajectory generation~\cite{liu2023aerialvln} marks progress toward this goal but does not leverage VLMs and relies on costly data collection and training procedures. In conclusion, a text-guided, VLM-based trajectory generation approach specifically designed for inspection planning remains a significant and unresolved research challenge.

\TE{Main Results:} As illustrated in~\prettyref{fig:teaser}, we propose the first training-free VLM-guided inspection planning algorithm for robot agents equipped with RGB camera sensors to inspect a series of POIs within given environments. The input to our algorithm consists of a 3D environmental map and a general text description that specifies the goal of a personalized inspection plan. This includes the set of POIs, the relative robot poses required to observe these POIs, and the desired order of POI visitation. To generate an inspection plan that aligns with the text description, our method is divided into three stages.
First, we construct a Probabilistic RoadMap (PRM), where the nodes represent a set of sampled observation positions. Next, we interact with the VLM to select a subset of nodes from which all POIs are visible from the desired robot poses. Finally, we solve a Traveling Salesman Problem (TSP) to determine the optimal visitation order of the POIs and perform trajectory optimization to compute a smooth and executable robot trajectory.
We evaluated our method across both handcrafted meshes and 3D scans of real-world environments. The results demonstrate that our approach can consistently generate trajectories that meet both the text description and the inspection-specific requirements, such as relative robot poses, trajectory smoothness, and/or time-optimality.
\begin{figure*}
\centering
\includegraphics[width=\linewidth]{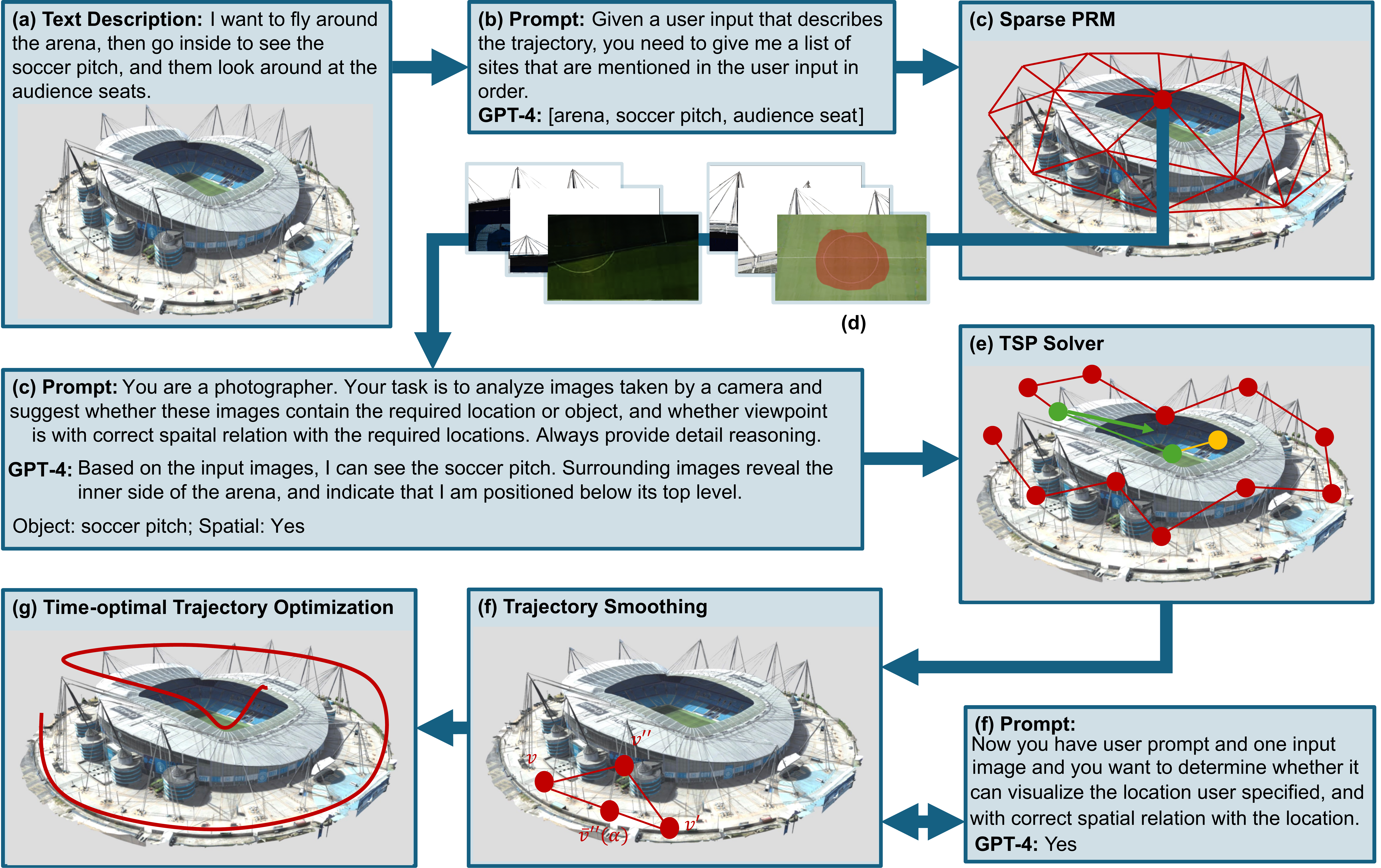}
\caption{\label{fig:pipeline}\footnotesize{The pipeline of our method. The input to our method is a text description and a 3D environmental map (a). Our method would first interact with VLM to extract a set of POIs $\mathcal{L}$ (b). Next, we construct a PRM and query each PRM node $v$ from which the POI $n_i$ is visible and salient, while also ensuring that the node satisfies optionally specified robot poses (c). We further ask the VLM to output a pixel-wise segmentation mask of $n_i$ from $v$, and treat the mass-center of the mask as the desired camera orientation focusing on $n_i$ (d). Next, we compute an initial trajectory connecting all the valid PRM nodes by solving a variant of TSP problems while preserving the optional visitation order of POIs (e), where different color indicates different POIs and the green arrow is the desired camera orientation. After the TSP problem provides an initial trajectory, we interact with VLM for iterative waypoint smoothing (f). Finally, we solve a continuous trajectory optimization, yielding the output (g).}}
\vskip -0.2in
\end{figure*}
\section{Related Work}

\noindent \paragraph{Inspection Planning} Early inspection planning pipelines decompose the task into two main steps: (i) selecting waypoints that collectively observe the set of POIs and (ii) generating a trajectory that passes through all waypoints while ensuring collision avoidance. Minimal-viewpoint solutions, based on Art-Gallery Problem~\cite{Danner2000RandomizedPF} and set-cover heuristics, provide coverage but do not guarantee an optimal overall plan. Thus, subsequent work has turned to variants of TSP~\cite{7765102}, trajectory optimization~\cite{8338068}, or iterative (re)-sampling of viewpoints to enhance path quality~\cite{7140101}.
More recent methods~\cite{7487440, almadhoun2016survey} leverage asymptotically optimal, sampling-based motion planners to couple viewpoint selection with trajectory generation. These approaches yield solutions whose cost converges to the optimum as the number of samples increases, albeit at a significant computational cost~\cite{gammell2021asymptotically, karaman2011sampling}. However, all of these methods require users to manually specify the set of POIs, limiting the ability to define tasks through more intuitive, high-level specifications such as text descriptions.
Motivated by the rise of VLMs, we introduce the first inspection planning framework that accepts free-form text commands. We use a VLM to localize semantic POIs and then invoke trajectory optimization to create a user-friendly, end-to-end inspection planning pipeline. The open-domain nature of VLMs further enables us to incorporate personalized constraints, such as relative robot poses for observing POIs and desired POI visitation orders, making the system more adaptable and intuitive.

\noindent \paragraph{LLM-guided Motion Planning} The success of pre-trained open-domain VLMs~\cite{radford2021learning} has significantly boosted the zero-shot performance of data-driven robotic systems in tasks such as table-top manipulation~\cite{liang2023code}, navigation~\cite{gu2022vision,zhou2024navgpt,sun2024beyond}, and embodied question answering~\cite{cheng2025efficienteqa}. These advancements reduce the high costs associated with data collection and fine-tuning when transferring to new problem domains. Recent results further confirm that Large Language Models (LLMs) can generate both coarse- and fine-grained motion plans~\cite{liang2023code}. In particular, \citewithauthor{hong20233d1, sun2026handle} demonstrated that multi-view images can help LLMs understand the 3D world, enabling them to perform planning tasks. These findings suggest that LLMs are capable of directly generating detailed robot agent inspection trajectories from text descriptions.
However, to the best of our knowledge, no existing work has leveraged VLMs specifically for inspection planning. One related effort, AerialVLN~\cite{liu2023aerialvln}, trained a VLN model using expert aerial agent trajectory data and crowd-sourced text annotations. However, their training data was not designed for inspection planning tasks. In contrast, our goal is to develop a zero-shot inspection planning system guided by open-domain VLMs, e.g., GPT-4 and Qwen-3-VL.
\section{Problem Statement \& Method}
The input to our method is a known 3D environmental map, denoted by $\mathcal{M}$, represented using either a mesh or a point cloud, as well as a user-specified text description, as illustrated in~\prettyref{fig:pipeline}, which contains a set of POIs and optionally the relative robot poses for observing these POIs, as well as the POI visitation order. The output of our method is the position trajectory $\tau_p(t):[0,T]\mapsto\mathbb{R}^3$ of an aerial or ground vehicle, with $T$ being the travel time. Our output trajectory not only respects the constraints specified by the text description, but is also efficient to execute with sufficient trajectory smoothness~\cite{mellinger2011minimum,sun2021fast}. 
We design a three-stage pipeline to automatically generate such trajectories guided by a VLM. As our main point of departure from prior works on VLN~\cite{gu2022vision} using sequential decision-making, our method needs to generate and optimize the entire trajectory all at once for efficient execution. Therefore, during our first stage, we extract the POIs and additional constraints by prompting the VLM. We then construct a PRM and detect the set of waypoints for observing each POI. During the second stage, we initialize and iteratively optimize the PRM-restricted trajectory for better smoothness under the constraint that the trajectory always conforms to the text description. Finally, we optimize a continuous trajectory using PRM nodes as waypoints to ensure efficient execution.

\subsection{Stage I: POI Extraction on PRM}\label{method:stage1}
We assume that \textit{``The text description consists of a set of POIs, with optional relative robot poses and visitation order constraints.''} Based on this assumption, we use the prompt shown in~\prettyref{fig:pipeline}b to extract two sets of POI descriptions, denoted as $\mathcal{L}^o\triangleq\left<n_1,n_2,\cdots,n_{L^o}\right>$ and $\mathcal{L}^u\triangleq\{n_{L^o+1},n_{L^o+2},\cdots,n_{L^o+L^u}\}$. $\mathcal{L}^o$ is an ordered set of $L^o$ POIs that must be visited sequentially. $\mathcal{L}^u$ is an unordered set of $L^u$ POIs that can be visited in any order. Each POI $n_i$ is represented as a pair of text descriptions, i.e., $n_i=\langle\text{name}_i,\text{pose}_i\rangle$, where $\text{name}_i$ is the name of POI and $\text{pose}_i$ is the optionally user-specified relative pose for the robot to observe the POI. In this work, we pre-specify the following set of possible spatial relations---$\langle\text{inside, over, in-front, around, arbitrary}\rangle$---where $\text{pose}_i$ is labeled as arbitrary if user did not specify any desired spatial relation. We append chain-of-thought in-context examples~\cite{wei2022chain} to reduce hallucination. Thanks to the reasoning capabilities of state-of-the-art VLMs, arbitrary user instructions can be processed. 

\begin{figure}[ht]
\centering
\includegraphics[width=0.9\columnwidth]{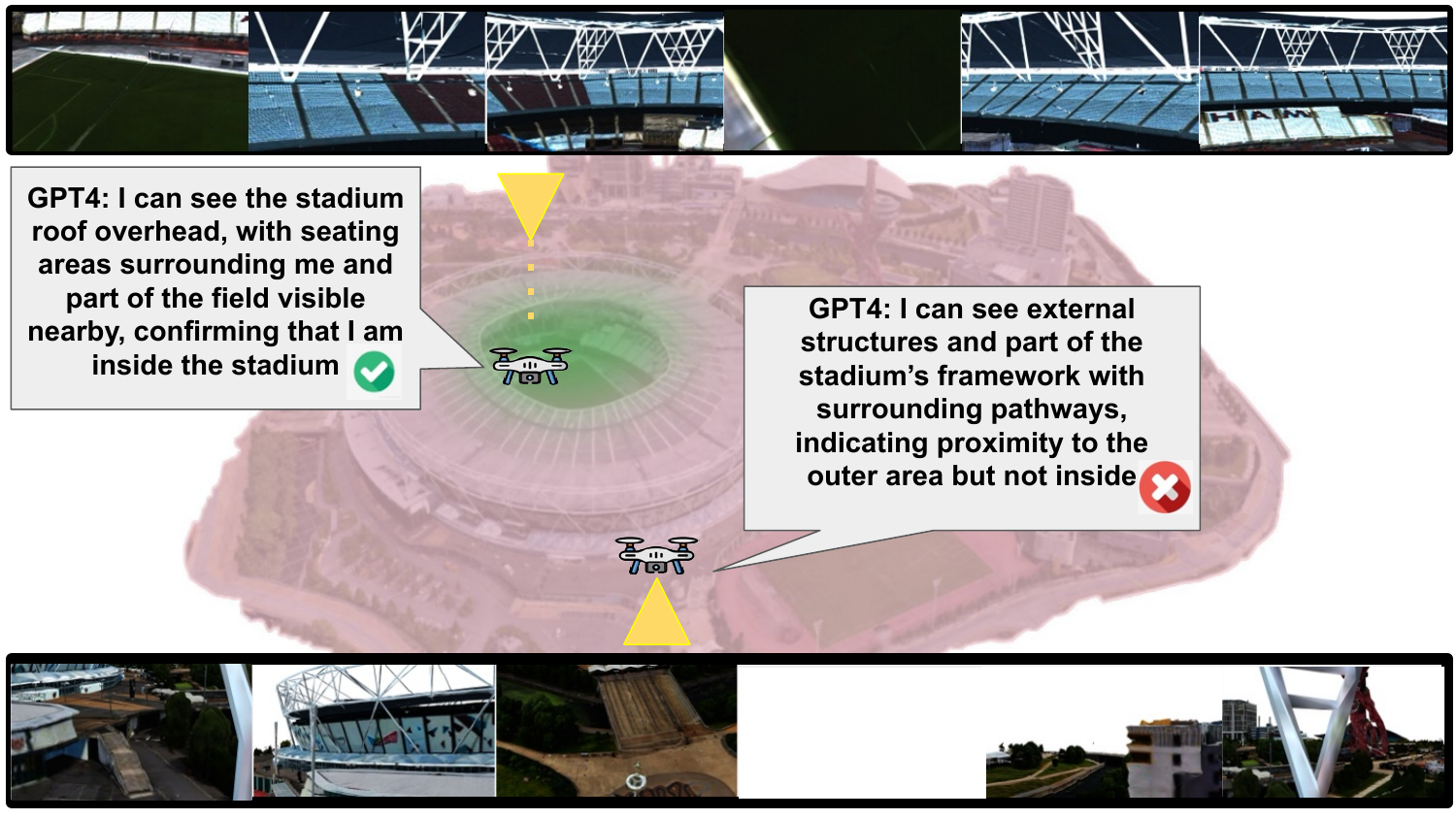}
\caption{\label{fig:spatial}\footnotesize{A heatmap visualization of the VLM's spatial understanding of the instruction \textit{``fly inside the stadium,"} where red and green indicates areas considered by VLM to be outside and inside the stadium, respectively. Six images are taken at every sample point and forwarded to the VLM. The six images from the two example views are shown on the top and bottom.}}
\vskip -0.3in
\end{figure}

For each POI $n_i$, we identify inspection positions from which it can be clearly observed. We construct a PRM for the free space of $\mathcal{M}$. Since na\"ive PRMs contain many nearby nodes with redundant visual information, leading to excessive VLM queries, we follow~\cite{6907541} and apply Poisson disk sampling. We first sample collision-free points, then sparsify them using Poisson disk sampling, enforcing a minimum spacing of $10\%$ of the scene bounding box. The resulting PRM is a graph $\mathcal{G}\triangleq\langle\mathcal{V},\mathcal{E}\rangle$ with node set $\mathcal{V}$ and edge set $\mathcal{E}$.
From each PRM node $v\in\mathcal{V}$, we capture six photos of the environment $\mathcal{M}$ from axis-aligned orientations ($\pm{X}, \pm{Y}, \pm{Z}$), collecting $6|\mathcal{V}|$ images. For each node, we prompt the VLM to evaluate whether any POI $n_i$ is visible from $v$ and whether the spatial relationship $\text{pose}_i$ is satisfied from $v$. The prompt includes only the user instruction, multi-view images, and chain-of-thought examples—no 3D model or camera poses. A potential pitfall is that VLMs assess visibility but not object saliency. To evaluate saliency, we propose to use an open-domain segmentation model, GroundedSAM~\cite{ren2024grounded}, which would predicts the mask of a POI from an image, as well as the Intersection over Union (IoU) score that approximately quantifies object saliency. Combining both models, we define the saliency indicator $\mathbb{I}_\text{salient}(n_i,v)$ where $\mathbb{I}_\text{salient}(n_i,v)=1$ if and only if the IoU of $n_i$ from any of the 6 images taken from $v$ is larger than 0.5 and the VLM confirms that $n_i$ is visible from $v$ and $v$ satisfies the specified $\text{pose}_i$, as shown in~\prettyref{fig:spatial}. If a photo contains multiple POIs, we simply duplicate the PRM node to generate exactly one node clone for each POI. After this procedure, we identify a set of potential waypoints for each $n_i$ denoted as $\mathcal{V}(n_i)=\{v\in\mathcal{V}|\mathbb{I}_\text{salient}(n_i,v)=1\}$, ensuring the inspection trajectory aligns with user commands.
\subsection{Stage II: PRM-restricted Trajectory Generation}\label{method:stage2}
We define a PRM node $v$ as valid if $v\in\mathcal{V}(n_i)$ for any $n_i$, and the set of valid nodes is denoted as $\bar{\mathcal{V}}\subseteq\mathcal{V}$. Note that the valid node set associated with different $n_i$ is also different due to node duplication. We insert an edge of length zero between a node and its duplications. For any pair of valid nodes $v$ and $v'$, their distance restricted to the PRM graph is $d(v,v')$. Specifically, we run the Dijkstra algorithm on the PRM graph to calculate the shortest path distance $d(v,v')$ for any $v, v' \in \bar{\mathcal{V}}$. Our first step is to select an as-smooth-as-possible trajectory restricted to the PRM graph such that each valid node is visited. We further require that the nodes are visited in the order specified in $\mathcal{L}_o$. In other words, for $i<j\leq L_o$, we ensure that nodes $v\in\mathcal{V}(n_i)$ appear before nodes $v'\in\mathcal{V}(n_j)$. Such a problem can be solved using an extension of TSP with additional order constraints. We formulate and solve such variant TSPs using Miller-Tucker-Zemlin (MTZ) method~\cite{desrochers1991improvements} to cast it as a mixed integer programming (MIP) as illustrated in~\prettyref{fig:pipeline}e. Specifically, we define binary auxiliary variable $x_{v\mapsto v'}\in\{0,1\}$, which takes value $1$ to indicate that the trajectory moves from $v$ to $v'$. Further, we define binary auxiliary variable $z_v\in\{0,1\}$ to indicate whether $v\in\bar{\mathcal{V}}$ is selected or not. We then solve the following mixed integer programming:
\small
\begin{align*}
&\argmin{\underset{z_v\in\{0,1\}}{\underset{x_{v\mapsto v'}\in\{0,1\}}{y_v\in[1,|\bar{\mathcal{V}}|-1]}}}\sum_v\sum_{v'\neq v}d(v,v')x_{v\mapsto v'}\\
&\ST
\begin{cases}
\sum_{v\in\mathcal{V}(n_i)}z_v\geq1\quad\forall n_i\\
\sum_{v'\neq v}x_{v\mapsto v'}=z_v\quad\forall v\\
\sum_{v\neq v'}x_{v\mapsto v'}=z_{v'}\quad\forall v'\\
y_{v'}-y_{v}\leq-1\quad\forall v\in\mathcal{V}(n_i), v'\in\mathcal{V}(n_j), i<j\leq L_o\\
y_{v'}-y_{v}+|\bar{\mathcal{V}}|x_{v\mapsto v'}\leq|\bar{\mathcal{V}}|-1\quad\forall v\neq v'
\end{cases}.
\end{align*}
\normalsize
The objective of the above MIP is to minimize the total travel distance, approximately maximizing smoothness. The first constraint ensures that there is at least one node observing each POI. The second and third constraints ensure that each selected node, as indicated by $z_v$, is visited exactly once, with one incoming and outgoing edge. The forth and fifth constraints ensure that the generated trajectory is connected and satisfies the specified visitation order constraint. Specifically, we introduce a continuous variable $y_v$ indicating the visitation order. The forth constraint ensures that, for a pair of nodes $v,v'$ observing two POIs $n_i,n_j$ with $i<j<L_o$, $n_j$ is visited later in the trajectory. The fifth constraint is the MTZ constraint, which ensures the trajectory is connected. Specifically, $y_v$ can be interpreted as the amount of objects stored in node $v$. The constraint ensures the robot agent removes one object when it visits each node. By restricting all $y_v$ in the range $[1,|\bar{\mathcal{V}}|-1]$, we essentially avoid disconnected trajectories. Finally, note that we do not care which valid node to visit first, so we introduce a dummy valid node as the first node, which is removed after the TSP solver. After this step, we arrive at an initial trajectory containing a set of waypoints $v\in\bar{\mathcal{V}}$. There can be other nodes $v''$ used by the Dijkstra algorithm to connect nodes in $\bar{\mathcal{V}}$.

The TSP trajectory can be wiggly and inefficient due to sparse PRM connectivity and POI constraints. Inspired by the Frank–Wolfe algorithm~\cite{nocedal1999numerical}, we propose an iterative VLM-guided trajectory smoothing scheme. We first perform brute-force simplification by replacing PRM-restricted segments with collision-free straight lines. We then examine intermediate nodes $v''$ between $v$ and $v'$, and try to smooth the trajectory by replacing $v''$ with the midpoint $\bar{v}''=(v+v')/2$ (\prettyref{fig:pipeline}f). Repeating this process gradually straightens the trajectory. However, we still need waypoint $v''\in\mathcal{V}(n_i)$ to observe the POI $n_i$ under consideration, and we need the POI to be visible with a high saliency after moving $v''$ to $\bar{v}''$. We achieve this using the VLM prompt in~\prettyref{fig:pipeline}f. If the VLM confirms saliency and spatial constraints and the motion is collision-free with $\mathcal{M}$, we update $v''$ to $\bar{v}''$. Otherwise we apply a backtracking line search during optimization. Specifically, we maintain a search step $\alpha$ and consider moving $v''$ to $\bar{v}''(\alpha)=\alpha\bar{v}''+(1-\alpha)v''$. We will iteratively query VLM at $\bar{v}''(\alpha)$. If the VLM confirms the saliency and spatial constraints are both preserved, we move $v''$ to $\bar{v}''(\alpha)$. Otherwise, we set $\alpha\gets\alpha/2$ and query again. The process stops when $\alpha<1/8$. Note that for nodes $v\notin\bar{\mathcal{V}}$, they are just used to connect other nodes and we can simply set them to midpoints without querying VLM.
The smoothing procedure continues until no nodes can be further improved. This process can be viewed as constrained optimization with the VLM acting as the constraint checker. However, repeated VLM queries may become the inference bottleneck. To reduce this cost, we prioritize nodes by curvature (the angle between edges $v''-v$ and $v''-v'$), processing the most curved first, avoiding repeated small updates with tiny $\alpha$. We report our runtime is relatively fast in~\ref{subsec: vlm_per}. This procedure is guaranteed to converge as it monotonically increases the smoothness of the trajectory.
\subsection{Stage III: Continuous Trajectory Generation}\label{method:stage3}
The smoothed trajectory from our last section is piecewise linear and inefficient for robot execution. Therefore, we design trajectory optimization solver based on~\cite{ni2021robust} to compute a collision-free spline trajectory (\prettyref{fig:pipeline}g). We first treat the nodes as control points and interpolate a composite B\`ezier curve. Note that such a B\`ezier curve might not be collision-free, although the node trajectory is. We therefore adaptively subdivide the nodes so the B\'ezier curve approximates the node trajectory. Subdivision continues until the curve is collision-free. We then solve the collision-free optimization of~\cite{ni2021robust}, minimizing integrated snap. We further introduce penalty terms to ensure that all the control points stay as close to the nodes as possible.
After optimization, node positions shift slightly due to soft penalties that keep them near the original PRM nodes. As observed in~\cite{ni2021robust}, these small adjustments preserve feasibility, maintaining spatial constraints and POI visibility. To further improve visibility, we query the VLM again to obtain POI masks with GroundingSAM and orient each camera toward the mask centroid. Optionally, camera orientation along the trajectory can be interpolated from the nodes using spherical barycentric coordinates~\cite{langer2006spherical}. At time instance $t$, the robot position is interpolated from nodes via the stencil:
\begin{align*}
\tau_p(t)=\sum_{j=1}^{d+1}b^j(t)v^j,
\end{align*}
where $v^j$ is the $j$th node, $b^j(t)$ is the Bernstein basis coefficient at time $t$, and $d$ is the order of the curve. Similarly, the camera-orientation is interpolated as:
\begin{align*}
\tau_o(t)=\text{Slerp}(b^1(t),\theta^1,\cdots,b^{d+1}(t),\theta^{d+1}),
\end{align*}
where $\theta^j$ is the camera direction, emanating from $v^j$ to the mass-center of the masked region for the POI, and finally Slerp($\bullet$) is the spherical barycentric interpolation function. We summarize our method using the pipeline~\prettyref{alg:pipeline}.

\small
\begin{algorithm}[ht]
\caption{VLM-guided Inspection Trajectory Generation}
\label{alg:pipeline}
\begin{algorithmic}[1]
\Require{Text description (\ref{fig:pipeline}a)}
\Ensure{$\tau_p,\tau_o$}
\State Extract POI list $\mathcal{L}_o$ and $\mathcal{L}_u$ (\ref{fig:pipeline}b)
\For{Each $v\in\mathcal{V}$}
\For{Each $i=1,\cdots,L_o+L_u$}
\State Compute $\mathbb{I}_\text{salient}(n_i,v)$ (\ref{fig:pipeline}c,~\ref{fig:pipeline}d)
\EndFor
\State Construct $\mathcal{V}(n_i)$
\EndFor
\State Solve MIP to compute initial node list (\ref{fig:pipeline}e)
\While{More nodes can be smoothed}
\State Order nodes in curvature-descending order
\For{Each consecutive $\left<v,v'',v'\right>$}
\State $\alpha\gets1$
\While{$\alpha\geq\underline{\alpha}$}
\If{$v''$ can be moved to $\bar{v}''(\alpha)$}
\State $v''\gets\bar{v}''(\alpha)$ (\ref{fig:pipeline}f)
\State Break
\Else
\State $\alpha\gets\alpha/2$
\EndIf
\EndWhile
\EndFor
\EndWhile
\State Solve trajectory optimization for $\tau_p$ (\ref{fig:pipeline}g)
\end{algorithmic}
\end{algorithm}
\normalsize

\begin{figure*}[t]
\centering
\includegraphics[width=0.32\linewidth]{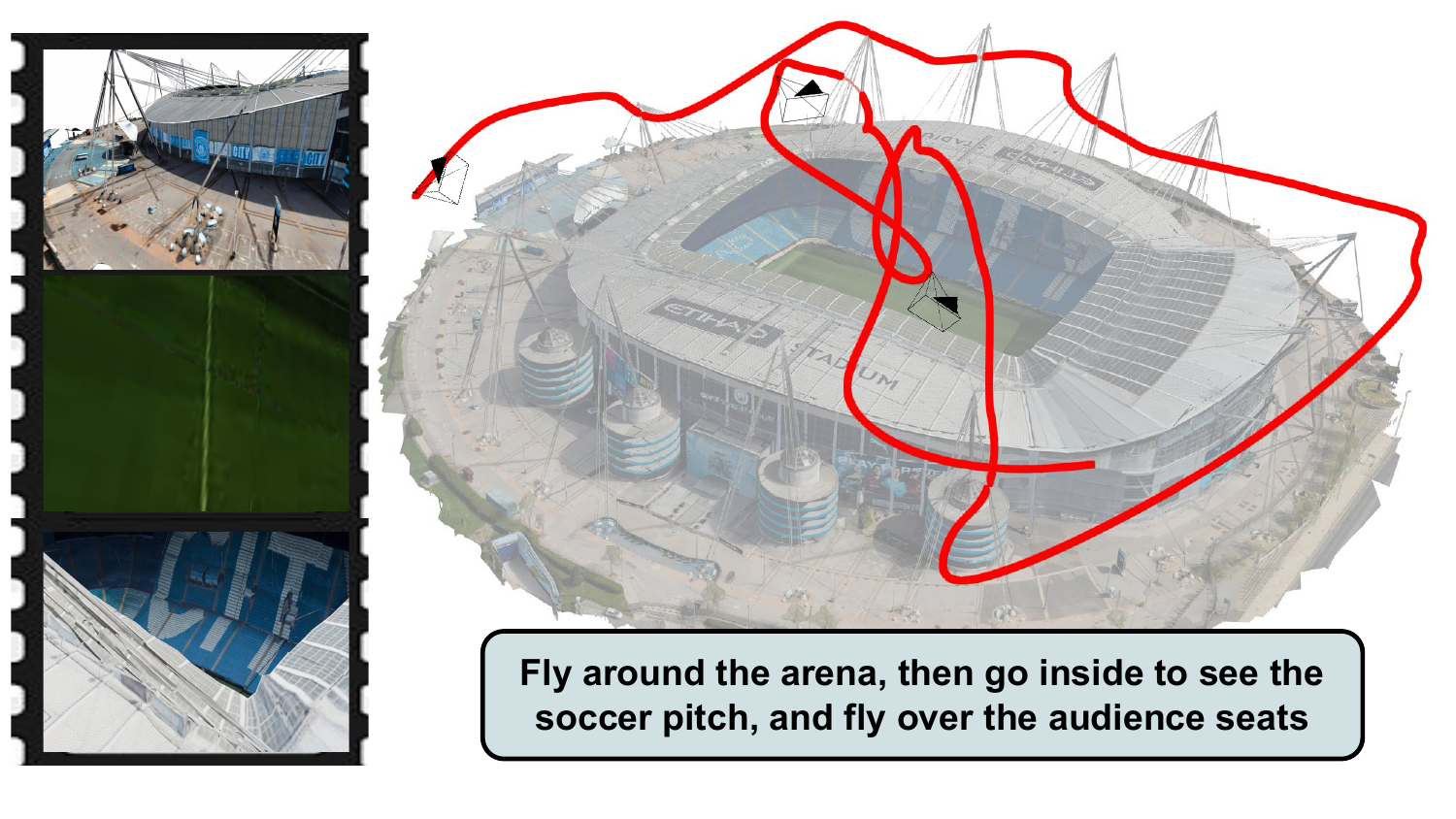}
\includegraphics[width=0.32\linewidth]{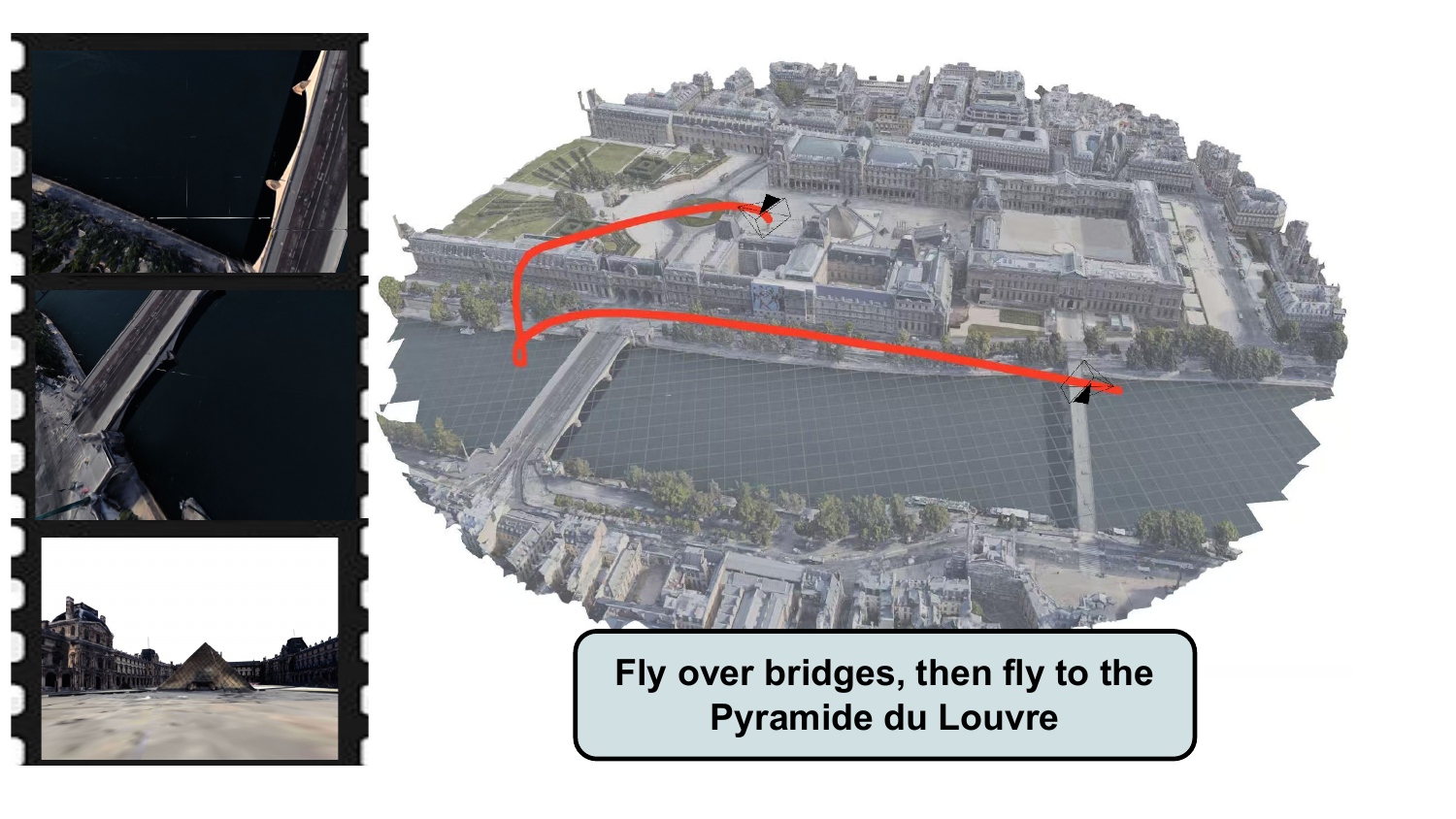}
\includegraphics[width=0.32\linewidth]{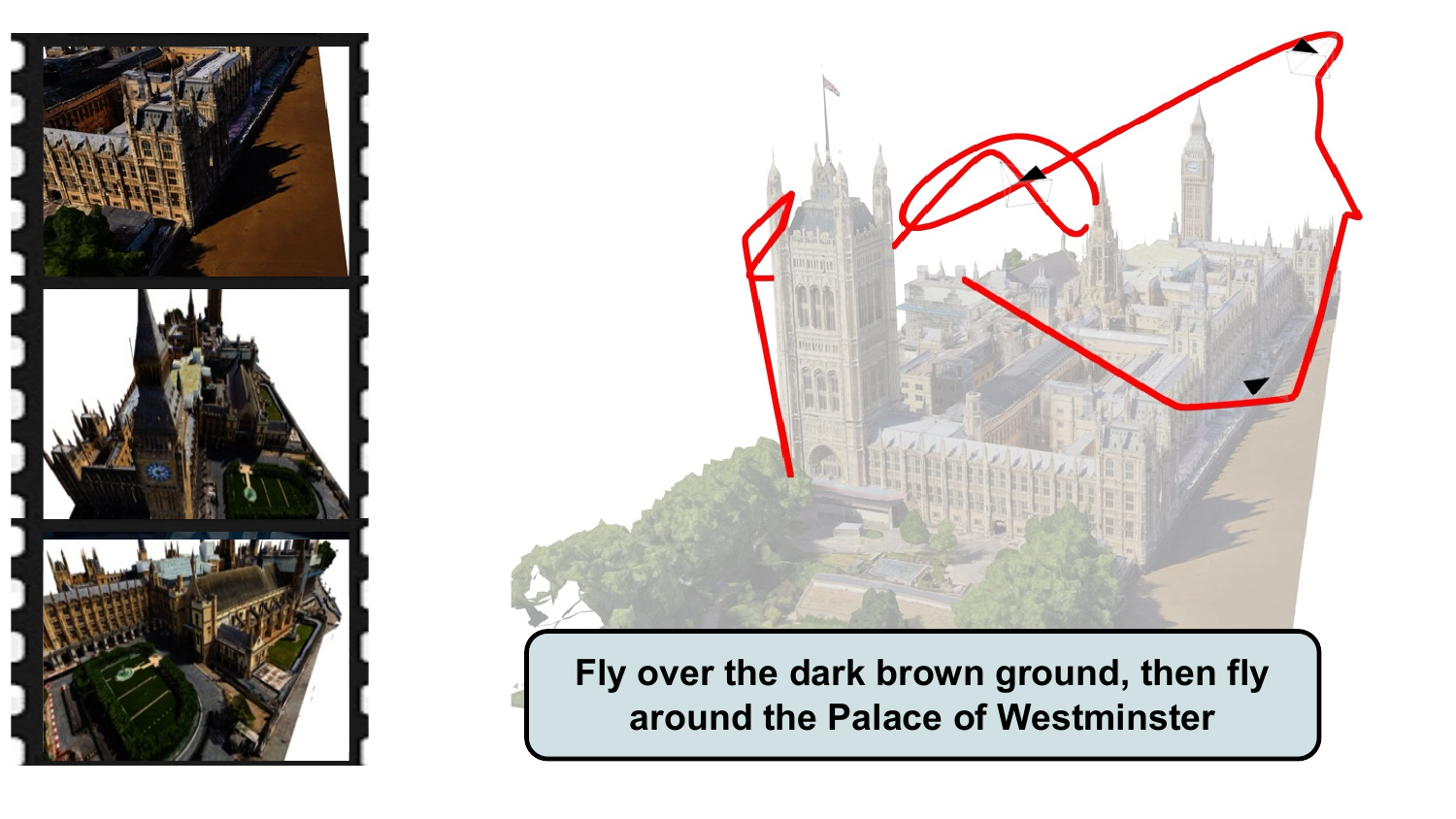}
\includegraphics[width=0.32\linewidth]{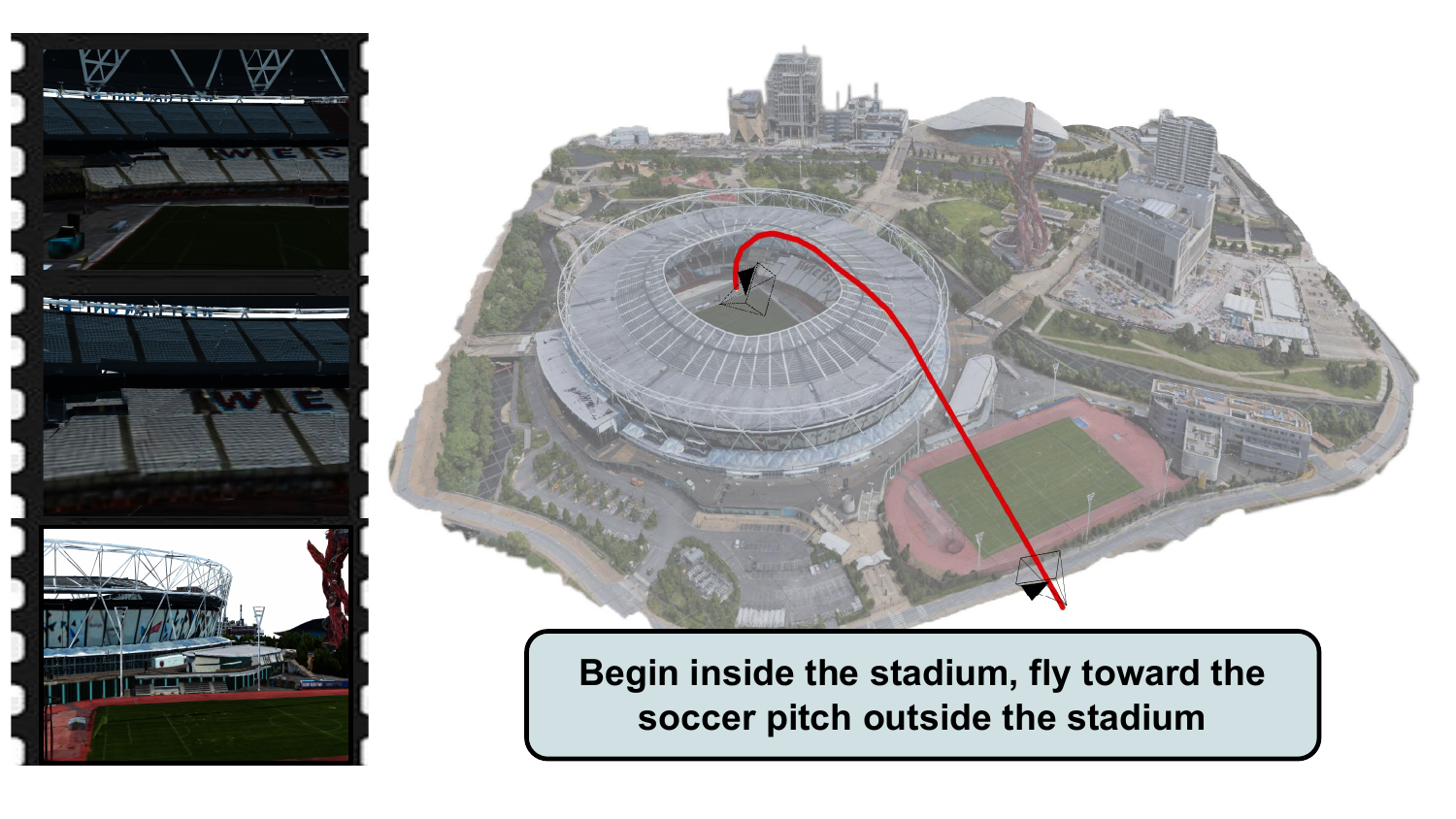}
\includegraphics[width=0.32\linewidth]{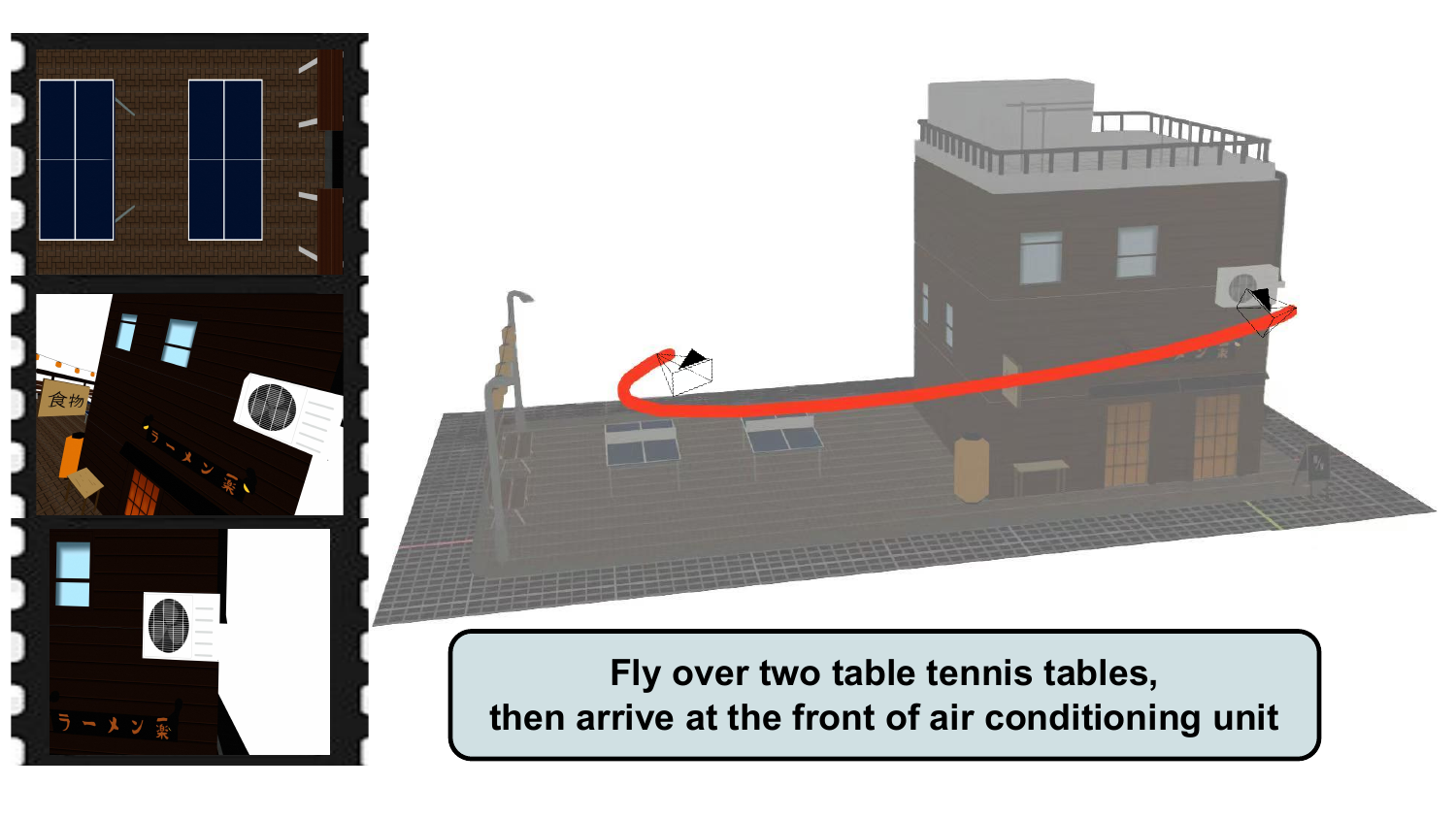}
\includegraphics[width=0.32\linewidth]{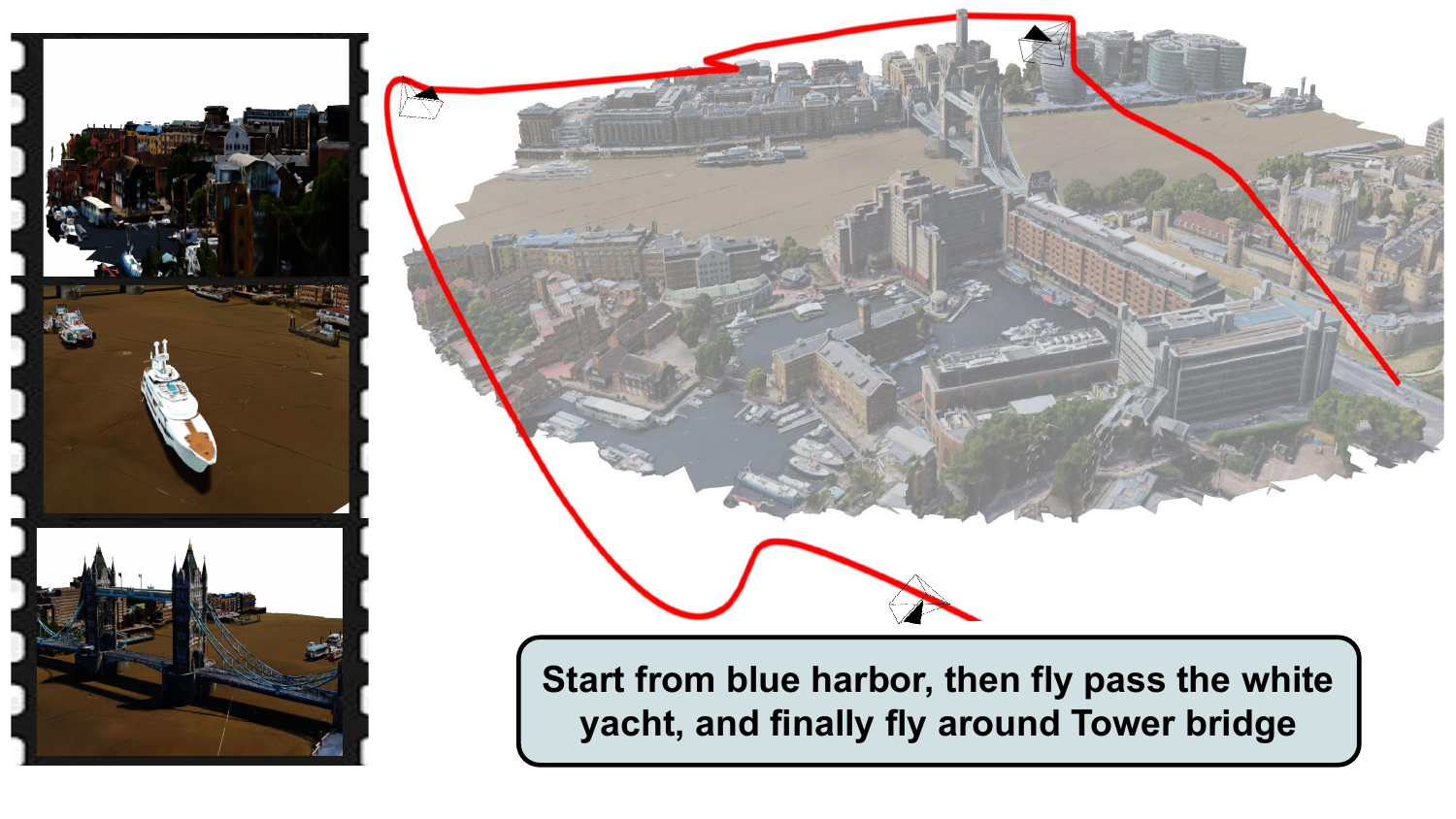}

\caption{\footnotesize{Outputs of our method. Inspection trajectories in red. Robot viewpoint camera frames of selected POIs are attached on the left side to highlight text conformity, with the related orientations marked along the trajectory. More visual comparison with baselines are shown in the supplemental video.}}
\label{fig:gallary}
\vskip -0.2in

\end{figure*}
\section{Evaluation}
\begin{figure}[thpb]
\centering
\includegraphics[width=0.49\linewidth]{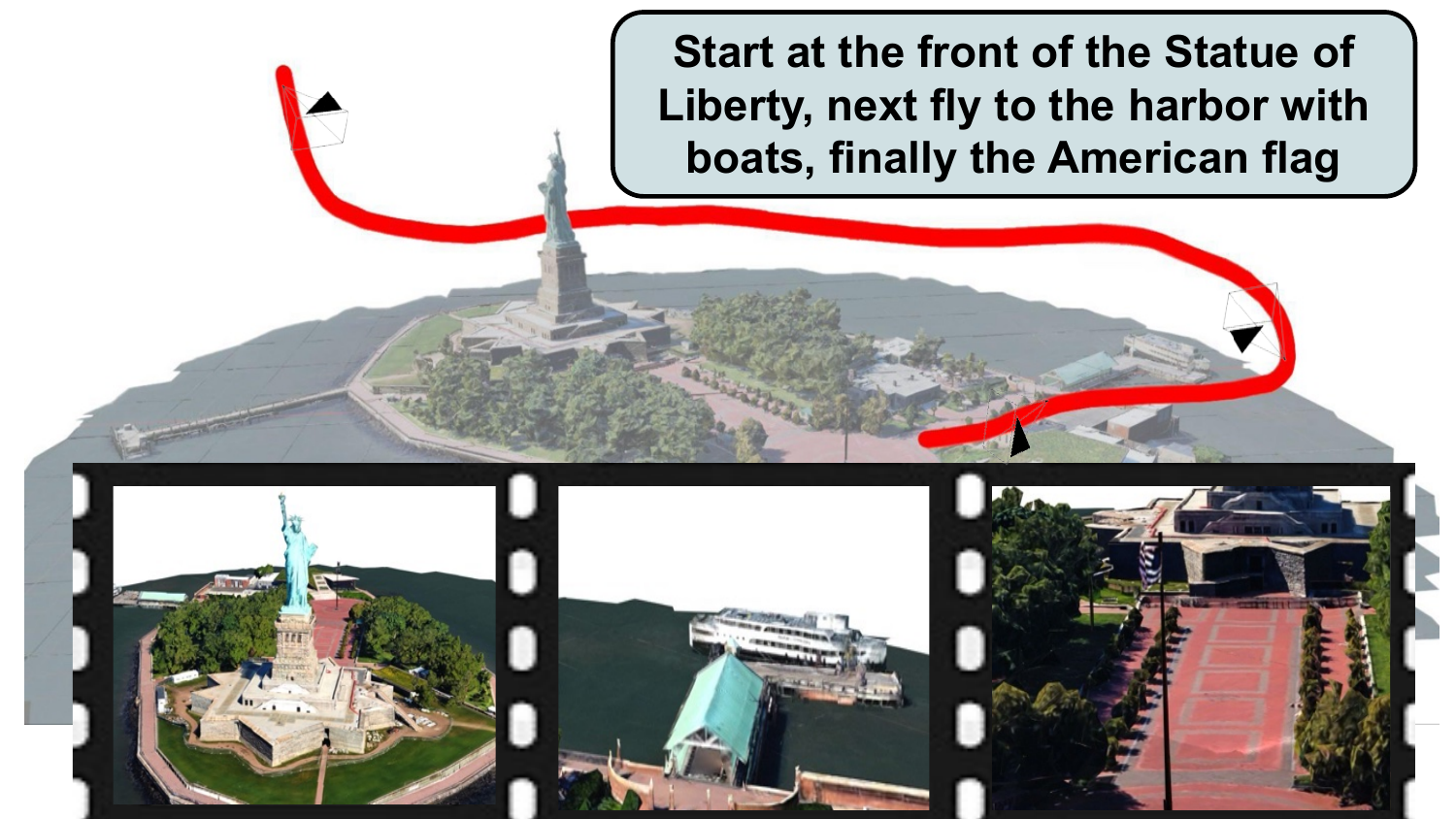}
\includegraphics[width=0.49\linewidth]{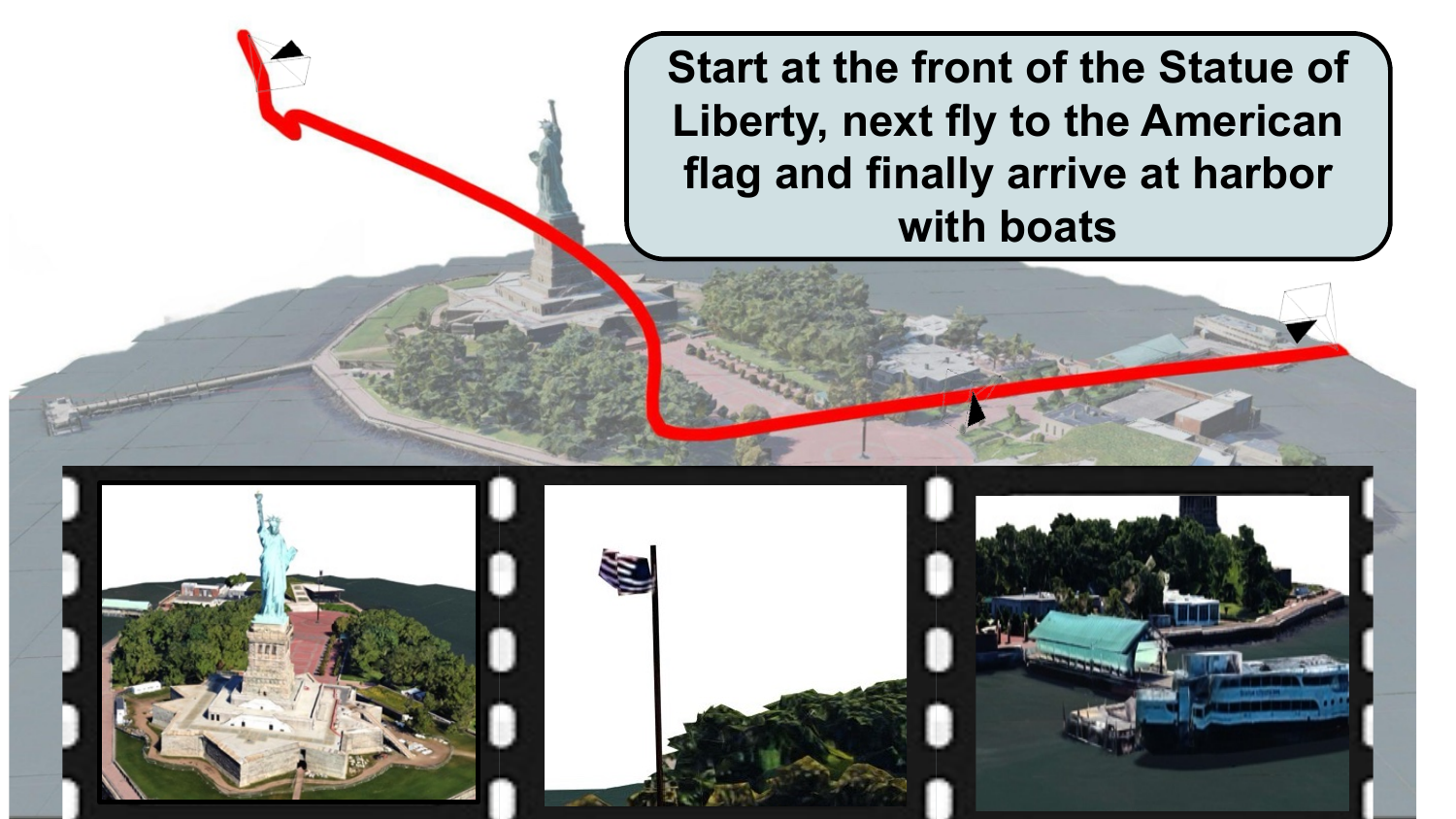}
\caption{\footnotesize{Results under different visitation order constraints. To follow personalized user text guidance, the left trajectory visits the ``harbor with boat" before the ``American flag," while the right trajectory visits the ``American flag" first.}}
\label{fig:visitation}
\vskip -0.2in
\end{figure}

We evaluate our method along two aspects of the generated trajectory: \textit{trajectory smoothness}, and \textit{text conformity} that measures the alignment between trajectory and text description. Experiments were run on a workstation with an RTX 3080 GPU. The VLM model used throughout is \textit{GPT-4o} with default parameters. Keypoint extraction uses one in-context example (\prettyref{fig:pipeline}b). Spatial relation reasoning uses eight in-context examples (\prettyref{fig:pipeline}c): one positive and one negative example for each of the four relations. The initial PRM contains 800 sampled nodes. Then in the Poisson disk sampling step, we ensured the sparsified graph $\mathcal{G}$ was fully connected to guarantee a solution in the subsequent TSP process, resulting in less node numbers VLM need to process in downstream thus to run pipeline efficiently. In~\prettyref{fig:gallary}, we show six trajectories generated using our method and highlight the key locations identified by VLM, where all the environments come from both scans of the real world and handcrafted meshes. In~\prettyref{fig:visitation}, we show that our pipeline conforms to personalized visitation order. Visually, our trajectory consistently agrees with the input text while being sufficiently smooth for efficient inspection. In~\prettyref{sec:real_uav}, we also conduct a real-world robot experiment to demonstrate our framework's effectiveness. Below, we conduct more detailed quantitative evaluations.

\subsection{VLM Performance}\label{subsec: vlm_per}
We evaluate the VLM's spatial reasoning capabilities based on multi-view images by measuring its winning rate in identifying spatial relationships and processing efficiency. We run 35 samplings for the 10 real-world scanned environments from Sketchfab website and calculate the average rate. To establish ground truth, we manually labeled the bounding boxes of key locations on 3D maps. The VLM's performance was assessed in determining whether sample points were \textit{``over'', ``inside'', ``in front'', or ``around''} key locations. \textit{``Over''} refers to points positioned on top of the bounding box, while \textit{``inside''} indicates points contained within it. \textit{``In front''} applies to points located in front of the key location, and \textit{``around''} includes points outside the bounding box but still within visible range of the key location.
We compare the spatial understanding performance of state-of-the-art VLMs, including GPT-4o, GPT-4o-mini, and the open-source Qwen2.5-VL-7B, using multi-view images, as summarized in~\prettyref{table:vlm-wr}. GPT-4o achieves over 90\% accuracy across all spatial relations, consistently outperforming other models, while Qwen2.5-VL-7B shows lower accuracy despite longer inference times. We suspect that the incorrect spatial reasoning is due to VLM halluciation.
We report the total VLM inference time for Stages~\prettyref{method:stage1} and~\prettyref{method:stage2}, averaged across all test cases. Despite the large environments (520k triangles and 305k vertices on average), our method runs within a practical runtime budget. We further confirm small-scale supervised fine-tuning (SFT) can improve VLM's spatial reasoning. Using a disjoint real-scanned environment also from Sketchfab, we collect 360 multi-view images from 60 datapoints and annotate then under 4 spatial relations. Consistent with official instruction on GPT websites, 50-100 examples can already improve GPT models, SFT leads to measurable gains in spatial reasoning accuracy (Table~\prettyref{table:vlm-wr}), leaving large-scale finetuning spatial reasoning based on multi-view images a possible future research direction. All code and datasets will be released upon acceptance.

\begin{table}[thpb]
\centering
\footnotesize
\caption{VLM Spatial Reasoning Win Rate and Run Time}
\label{table:vlm-wr}
\begin{tabular}{l|c|cccc}
\toprule
\textbf{Model} & \textbf{Time (s)} & \multicolumn{4}{c}{\textbf{Winning Rate (\%)}}\\
 & & \textit{Over} & \textit{Inside} & \textit{In Front} & \textit{Around} \\
\midrule
GPT-4o       & 96  & 96.21 & 91.07 & 92.36 & 94.58 \\
\quad + SFT & 96 & 97.84 & 93.52 & 94.91 & 96.13 \\
\midrule
GPT-4o-mini  & 91  & 91.12 & 87.03 & 89.42 & 92.87 \\
\quad + SFT & 91 & 93.67 & 89.94 & 91.76 & 94.25 \\
\midrule
Qwen2.5{\scriptsize-VL-7B} & 101 & 83.74 & 78.66 & 80.12 & 84.91 \\
\quad + SFT & 101 & 88.45 & 84.12 & 85.73 & 89.06 \\
\bottomrule
\end{tabular}
\vskip -0.15in
\end{table}

\begin{table}[thpb]
\centering
\footnotesize
\caption{Ablation study on trajectory smoothness}
\label{table:ablation}
\begin{tabular}{l|ccccc}
\toprule
Method & $\kappa$ & $J$ & steps & distance & CLIP \\
\midrule
Ours & \textbf{0.39}& \textbf{0.0036}& \textbf{26.75}&\textbf{2612.85}&\textbf{0.2959}\\
\midrule
-Smoothness& 0.43& 0.0053& 26.75 &2748.30& 0.2897\\
-Simplify-Smoothness& 0.46& 0.0102& 37.67& 3825.60& 0.2814\\
-Low Resolution& 0.40& 0.0042& 28.10& 2668.40& 0.2931\\
\bottomrule
\end{tabular}
\vskip -0.25in
\end{table}

\subsection{Smoothness}
The smoothness of the generated trajectory is evaluated using two key metrics: curvature and jerk. In general, lower curvature and jerk value indicate better smoothness of the trajectory. Curvature, denoted as $\kappa$, describes how sharply a curve bends at a particular point. The curvature $\kappa$ of a trajectory at a point is defined as: $\kappa\triangleq\|\dot{\tau}_p\times\ddot{\tau}_p\|/\|\dot{\tau}_p\|^3$. Here $\dot{\tau}_p$ is the velocity vector and $\ddot{\tau}_p$ is the acceleration vector. Jerk represents the rate of change of acceleration over time and indicates the smoothness of motion transitions. For a 3D trajectory, the jerk is defined as the magnitude of the third derivative of the position vector with respect to time $t$, defined as: $J\triangleq\int_0^T\|\dddot{\tau}_p(t)\|dt$. We further introduce two more metrics: i) Steps are defined as the number of line-segment pieces in the PRM-restricted trajectory, i.e., the trajectory before using the trajectory optimization solver in \prettyref{method:stage3}; ii) Distance is defined as the total length of our generated trajectory. 
We evaluate our method against 5 baselines: PRM+VLM, LM-Nav~\cite{shah2023lm}, VoroNav\cite{wu2024voronav},
UniGoal\cite{Yin_2025_CVPR}, and Oracle. For PRM+VLM, we construct a Poisson-disk-sampled PRM, use a VLM to identify nodes corresponding to key locations, and connect them sequentially in random order—effectively applying only Stage~I of our framework. LM-Nav~\cite{shah2023lm}, VoroNav\cite{wu2024voronav}, and
UniGoal\cite{Yin_2025_CVPR} are zero-shot navigation method that generates paths from high-level text instructions using different topological graph with front-view only images but does not account for spatial constraints, smoothness, or orientation planning, which our approach explicitly addresses. Oracle represents a smooth trajectory formed by connecting user-selected waypoints from each sampling that satisfy the POI and spatial requirements, serving as an upper bound on the achievable trajectory. 
We report averaged metrics over all text prompts and mesh environments shown in~\prettyref{fig:gallary}, summarized in~\prettyref{table:baseline}. Our method achieves superior $\kappa$, steps, distance metrics, and is closest to the Oracle upper bound. To further validate our design choices, we conduct an ablation study (\prettyref{table:ablation}) by removing (1) the VLM-guided smoothing step (–Smoothness), (2) both smoothing and path simplification (–Simplify-Smoothness), and (3) downscaling the original 1920×1080 rendered images by 0.5× to simulate lower-resolution imagery typical of drone-mounted cameras (–Low Resolution). Our full method outperforms both ablated variants across all metrics, confirming the effectiveness of each component. Performance also remains largely invariant to image resolution: even with downsampled inputs, our pipeline still generates collision-free trajectories that visit all required POIs with 100\% correct spatial relations, closely matching the high-resolution results, due to the multi-scale robustness of VLMs and SAM.

\begin{table}[thpb]
\centering
\footnotesize
\caption{Comparison of trajectory smoothness, efficiency, and CLIP alignment}
\label{table:baseline}
\begin{tabular}{l|ccc|cc}
\toprule
\multirow{2}{*}{Method} & \multirow{2}{*}{$\kappa$} & \multirow{2}{*}{Steps} & \multirow{2}{*}{Distance} & \multicolumn{2}{c}{CLIP} \\
 & & & & ViT-B/32 & ViT-B/16 \\
\midrule
Oracle & 0.3728 & 24.00 & 2448.63 & 0.2911 & 0.2984 \\
PRM+VLM & 0.4834 & 488.25 & 47192.89 & 0.2719 & 0.2716 \\
LM-Nav\cite{shah2023lm} & 0.4769 & 87 & 10544.30 & 0.2726 & 0.2714 \\
VoroNav\cite{wu2024voronav} & 0.4521 & 54.50 & 6821.34 & 0.2768 & 0.2795 \\
UniGoal\cite{Yin_2025_CVPR} & 0.4589 & 41.50 & 5342.77 & 0.2781 & 0.2762 \\
Ours & \textbf{0.3865} & \textbf{26.75} & \textbf{2612.85} & \textbf{0.2897} & \textbf{0.2959} \\
\bottomrule
\end{tabular}
\vskip -0.25in
\end{table}


%

\subsection{Text Conformity}
The text conformity of the generated trajectory is assessed using the CLIP score~\cite{hessel2021clipscore}, which quantifies the alignment between the visual output and the text description of the trajectory. The CLIP score is defined as: $
\text{CLIP} = \frac{F_{\text{image}} \cdot F_{\text{text}}}{\|F_{\text{image}}\| \|F_{\text{text}}\|},
$
where $F_{\text{image}}$, and $F_{\text{text}}$ are the feature vectors of the image and text produced by the CLIP model, respectively. The CLIP score ranges from -1 to 1, with higher values indicating better conformity to the user-specified locations. For each trajectory, we compute the maximum CLIP score across all camera views relative to each user-specified location. As shown in~\prettyref{table:baseline}, we report the average of these maximum scores under two CLIP vision transformers, ViT-B/32 and ViT-B/16~\cite{radford2021learning}. Our method achieves comparable CLIP scores to the baselines, demonstrating that the trajectory smoothing and time-optimal optimization stages~(\prettyref{method:stage2},~\prettyref{method:stage3}) preserve text conformity. Additionally, reorienting the camera toward the mass center of each target improves alignment, yielding a $0.02$ increase in CLIP score.
Text conformity and inspection quality is further evaluated using the mean IoU score across all keypoint images to quantify how well the VLM-selected camera views align with the text prompt. GPT-4o, GPT-4o-mini, and Qwen2.5-VL-7B achieve IoU scores of 0.907, 0.852, and 0.805, respectively. While performance slightly decreases for smaller models, the results demonstrate the robustness of our framework in generating high text-conforming trajectories, and having location inspected with high quality across the flight.

\subsection{VLM Query Scalability and Acceleration}
In our method, the most VLM queries are used to identify the set of inspection positions from PRM nodes. Therefore, the number of VLM queries scales linearly with the number of POIs. For each query, we take six pictures (render time is negligible) from each node and generate one query for each POI. To reduce the computational time for processing multiple images, we combine six sub-images into a single composite image, each labeled with its corresponding spatial information. On average, the system makes 89.12 VLM API calls per run, with a minimum of 70 and a maximum of 120. We further employ asynchronous programming to send multiple API requests in parallel, which significantly improves throughput when handling large batches of prompts by maximizing the tokens-per-minute (TPM) utilization.

\subsection{POI Coverage}
As a crucial criterion for a robust inspection planner, the generated inspection trajectory should cover each and every POI mentioned in the text description. Our evaluation of POI coverage is divided into two parts. First, we check whether the VLM can successfully extract all the POIs from the text description. The success rate of POI extraction and camera view identification is $100\%$ throughout all our benchmarks and all three VLMs: GPT-4o, GPT-4o-mini, and Qwen2.5-VL-7B. Second, with the extracted POI, we need to ensure the set of camera views generated using PRM can cover all the POIs. To evaluate POI coverage, we profile the number of triangles in the 3D mesh that can be observed by at least one of the camera view. Over all the benchmarks, the mean/minimum/maximum fraction of visible triangles is 76.92\% / 68.57\% / 82.36\%. Such a high coverage justify that the number of PRM nodes we choose is appropriate to achieve a high success rate to cover all POIs, without an excessive number of VLM queries. In our benchmark set, the VLM achieve 100\% success rate in identifying camera views containing all POIs. 

\begin{figure}[thpb]
\centering
\includegraphics[
width=0.98\linewidth,
trim=0 3cm 0 1.2cm,
clip
]{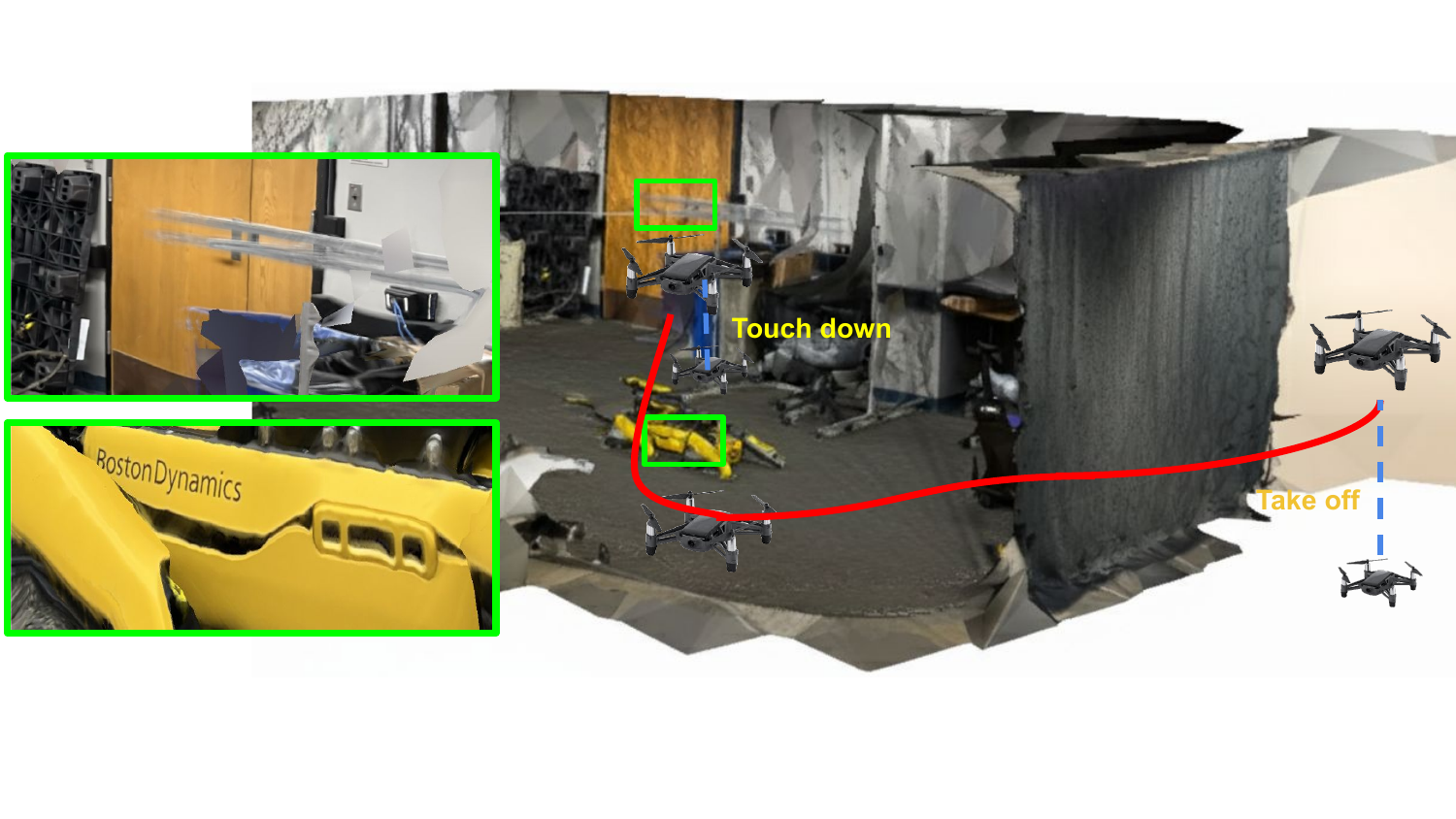}
\caption{\footnotesize{Real-world demonstration of the instruction: ``Fly to check that the \textit{Boston Dynamics} robot’s logo at the center of the room is clearly visible, and confirm that the door is closed.'' The DJI Tello drone follows the planned trajectory generated from the reconstructed 3D environment.}}
\label{fig:real-world}
\vskip -0.2in
\end{figure}

\subsection{Real-World Experiment}\label{sec:real_uav}
We also validate our framework through real-world experiments using a DJI Tello drone. We first scan the environment as a 3D mesh using RealityScan mobile application. The planned trajectory is then computed offline on a workstation and calibrated to the drone's coordinate frame via the DJITelloPy library. Communication with the drone is established over a 2.4\,GHz Wi-Fi link, while video transmission operates at 5.8\,GHz. The reconstructed environment and the investigation task are shown in~\prettyref{fig:real-world}. The drone is tasked to ensure that the lab environment is secure and ready for a quadruped robot environment, where we request the drone to check that the correct-type of quadruped robot is used and the door is closed. The drone takes off from its starting position and lands after completing the inspection flight.

\section{Conclusion \& Limitations}
We propose the first VLM-guided inspection trajectory generation method from general text descriptions. We identify several limitations for further exploration. First, due to the trajectory-wise nature of our operations, our methods need a large number of VLM queries with vision inputs, leading to a high inference cost. In the future, we consider more efficient ways to solve the VLM-constrained trajectory optimization problems. A potential solution is to have VLM generate a constraint that can be handled using conventional numerical optimization~\cite{tsouros2023holy}. Although we integrate GroundingSAM [31] to assist the VLM in reasoning about visual input, the VLM may still experience hallucinations [40], which can affect our method’s reliability,
particularly when it hallucinates and fails to identify a spatial
constraint. Fine-tuning the VLM with a large multi-view image dataset is a possible way to improve its spatial relation understanding. Finally, our current solution requires remote VLM inferences on remote machines, which would cause potential privacy issues, which could be mitigated by performing local VLM inferences. We could even consider performing different VLM queries on different models. For example, the POI extraction only requires text-based inference and does not involve visual information, which could be performed on a smaller model via local inference. In this way, the user-input text description is invisible to the remote machines.

\AtNextBibliography{\footnotesize}
\printbibliography

\end{document}